\documentclass[conference]{IEEEtran}
\IEEEoverridecommandlockouts
\usepackage{booktabs}
\usepackage{cite}
\usepackage{amsmath,amssymb,amsfonts}
\usepackage{graphicx}
\usepackage{textcomp}
\usepackage{hyperref}
\def\BibTeX{{\rm B\kern-.05em{\sc i\kern-.025em b}\kern-.08em
    T\kern-.1667em\lower.7ex\hbox{E}\kern-.125emX}}
\usepackage{balance}

\hypersetup{hidelinks}

\usepackage{eso-pic}
\newcommand\AtPageUpperMyleft[1]{\AtPageUpperLeft{
 \put(\LenToUnit{1cm},\LenToUnit{-1cm}){
     \parbox{\textwidth}{\raggedright\fontsize{9}{11}\selectfont #1}}
 }}
\newcommand{\conf}[1]{
\AddToShipoutPictureBG*{
\AtPageUpperMyleft{#1}
}
}

\begin{document}

\title{SkySeaLand: A Wide-Format Satellite Transportation Benchmark with an Ultra-Lightweight Detection Baseline}
\conf{This work has been submitted to the IEEE for possible publication. Copyright may be transferred without notice, after which this version may no longer be accessible.}

\author{\IEEEauthorblockN{
Md. Zahid Hasan Riad\IEEEauthorrefmark{1} and
Md Sultanul Islam Ovi\IEEEauthorrefmark{2}
}
\IEEEauthorblockA{\IEEEauthorrefmark{1}Dept. of Computer Science and Engineering, Green University of Bangladesh, Bangladesh}
\IEEEauthorblockA{\IEEEauthorrefmark{2}Dept. of Computer Science, George Mason University, Virginia, USA}
\IEEEauthorblockA{Email: 201002313@student.green.ac.bd, movi@gmu.edu}
}

\maketitle

\begin{abstract}
Satellite object detection is challenged by small targets and wide-format
scenes that lose detail under standard square-input resizing. We introduce
SkySeaLand, a public dataset of 1,307 high-resolution satellite images and
19,101 verified bounding boxes across airplane, boat, car, and ship classes
in terrestrial and maritime scenes. Native COCO and YOLO annotations are
provided. The collection is dominated by large source images and wide scene
geometry: 84.5 percent exceed 3,836 pixels on the longest side and 73.1
percent are near a 3:1 aspect ratio. We evaluate twelve detectors from the
YOLO, RT-DETR, DETR, and Faster R-CNN families using a common split and COCO
metrics. The tested YOLO and RT-DETR variants obtain 84.4--88.2 mAP50, with
no consistent accuracy gain from larger parameter counts under the reported
model-specific recipes. We also report SkyDet, a 1.22 M parameter
anchor-free baseline that obtains 60.5 mAP50 and 24.32 mAP50-95 in a 4.90 MB
footprint, with 13.74 ms latency (72.8 FPS) on a Tesla T4. SkySeaLand provides
a compact benchmark for mixed land--maritime transportation detection, while
SkyDet establishes a documented low-footprint reference rather than a
state-of-the-art accuracy claim.
\end{abstract}


\section{Introduction}

Object detection in satellite imagery supports traffic analysis, port activity assessment, airport surveillance, and maritime search \cite{cheng2016survey}. The problem differs from natural-image detection in several ways. Objects of interest occupy a small fraction of the scene, often below $32 \times 32$ pixels \cite{lin2014microsoft}. Scenes are acquired at high resolution and frequently in wide formats that follow runways, coastlines, and shipping lanes rather than the near-square frames common in ground-level photography \cite{xia2018dota}. A detector built for satellite use must therefore handle large scale variation and non-standard image geometry together.

Public benchmarks have driven progress in this area. DOTA \cite{xia2018dota}, DIOR \cite{li2020object}, NWPU VHR-10 \cite{cheng2014multi}, and xView \cite{lam2018xview} established evaluation protocols for aerial detection, and detector families originally developed on MS COCO \cite{lin2014microsoft}, such as Faster R-CNN \cite{ren2015faster}, the YOLO series \cite{redmon2016you,khanam2024yolov11,tian2025yolov12,jocher2026yolo26}, and DETR-style transformers \cite{carion2020end,zhao2024detrs}, have been adapted to them. Two practical gaps remain. Existing aerial benchmarks rarely combine terrestrial and maritime transportation classes in one label space while releasing both COCO and YOLO annotations. In addition, the sub-2 M parameter regime is seldom included in comparative aerial-detector tables, leaving a useful low-footprint reference point uncharacterized.

This paper addresses both gaps. We introduce SkySeaLand, a curated dataset of 1,307 satellite images with 19,101 verified bounding boxes over four classes: airplane, boat, car, and ship. Images cover airports, highways, harbors, marinas, and coastal regions across several geographic regions, and annotations are released in both COCO and YOLO formats. The source images are predominantly large (84.5\% above 3,836 px on the longest side) and wide (73.1\% near a 3:1 aspect ratio), stressing the information loss imposed by square-input resizing. We benchmark twelve detectors spanning four architectural families and report SkyDet, an anchor-free detector with 1.22 M parameters and a 4.90 MB checkpoint, as an explicitly low-footprint reference. The contributions are:

\begin{figure*}[!htbp]
\centering
\includegraphics[width=.9\textwidth]{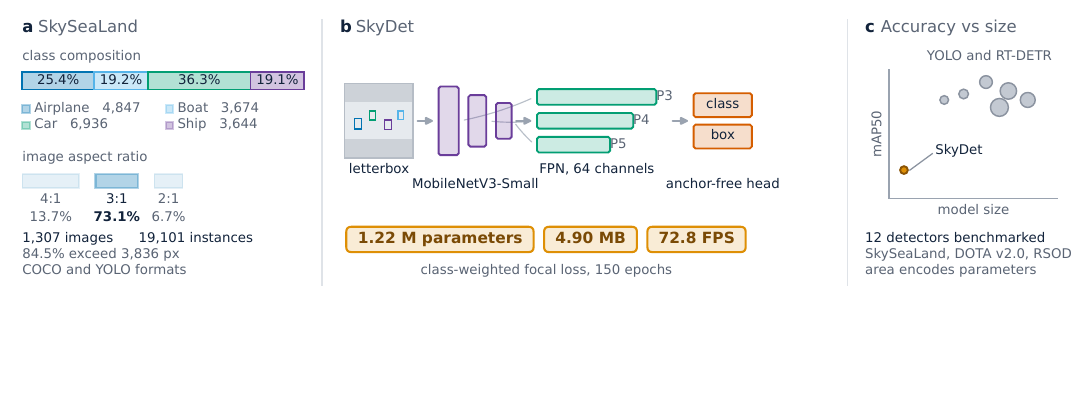}
\caption{Overview of SkySeaLand and SkyDet. (a) Class composition and aspect-ratio distribution across 1,307 images and 19,101 annotated instances. (b) SkyDet pairs a MobileNetV3-Small backbone with a 64-channel feature pyramid and an anchor-free detection head. (c) Among twelve benchmarked detectors, SkyDet occupies a distinct low-footprint operating point.}
\label{fig1}
\end{figure*}

\begin{enumerate}
\item \textbf{SkySeaLand}, a publicly released satellite object detection dataset (Kaggle and Mendeley Data) covering four transportation classes across land and maritime scenes, with dual-format annotations and a documented geometry profile.
\item \textbf{A twelve-detector practical benchmark} spanning YOLO (v10/v11/v12/v26), RT-DETR, DETR, and Faster R-CNN, with accuracy, parameter count, checkpoint size, and hardware-qualified latency.
\item \textbf{A documented low-footprint baseline}, SkyDet, built from a MobileNetV3-Small backbone \cite{howard2019searching}, a 64-channel feature pyramid \cite{lin2017feature}, and an anchor-free head. It retains 69\% of YOLOv11m's mAP50 with 6\% of its parameters and runs at 13.74 ms per image on a Tesla T4.
\end{enumerate}

\section{Related Work}

\textbf{Aerial and satellite detection datasets.} DOTA \cite{xia2018dota} contains 2,806 large images and over 188k instances across 15 classes (18 in v2.0). DIOR \cite{li2020object} provides 23,463 images over 20 classes at a fixed $800 \times 800$ resolution. NWPU VHR-10 \cite{cheng2014multi} is an earlier 10-class benchmark with 800 images. RSOD \cite{long2017accurate} covers four classes in 976 images. xView \cite{lam2018xview} scales to one million instances over 60 classes from WorldView-3 imagery, while VisDrone \cite{zhu2021detection} targets drone-altitude scenes with dense small objects. AI-TOD \cite{Wang2021} isolates the tiny-object regime, with a mean object size of 12.8 pixels.

\begin{table}[!t]
\caption{Context Among Aerial Object Detection Datasets}
\label{tab:dataset_context}
\centering
\scriptsize
\setlength{\tabcolsep}{2pt}
\begin{tabular}{@{}lrrcll@{}}
\toprule
\textbf{Dataset} & \textbf{Images} & \textbf{Cls.} & \textbf{Box} &
\textbf{Scope / L--M} & \textbf{Native labels} \\
\midrule
DOTA v1.0 & 2,806 & 15 & OBB & General / subset & DOTA TXT \\
DIOR & 23,463 & 20 & HBB & General / subset & VOC XML \\
NWPU VHR-10 & 800 & 10 & HBB & General / subset & TXT \\
RSOD & 976 & 4 & HBB & Four-class / no & VOC XML \\
xView & 1,129 & 60 & HBB & Broad / subset & GeoJSON \\
\textbf{SkySeaLand} & 1,307 & 4 & HBB & Transport / dedicated & COCO+YOLO \\
\bottomrule
\end{tabular}
\vspace{1pt}

\parbox{\columnwidth}{\scriptsize L--M: land--maritime transportation coverage. ``Subset'' denotes relevant classes within a broader ontology; formats refer to original releases.}
\end{table}

Table~\ref{tab:dataset_context} makes the scope distinction explicit. Several broader benchmarks contain both land and maritime transportation categories, but only as subsets of heterogeneous ontologies. SkySeaLand dedicates its complete four-class label space to that mixed transportation setting, releases ready-to-use COCO and YOLO annotations, and preserves predominantly wide, untiled source images. At 1,307 images, it is a compact complementary benchmark rather than a replacement for larger corpora.

\textbf{Object detectors.} Faster R-CNN \cite{ren2015faster} introduced learned region proposals, and feature pyramid networks \cite{lin2017feature} established multi-scale feature fusion. One-stage detectors traded proposals for speed, beginning with YOLO \cite{redmon2016you} and SSD \cite{liu2016ssd}; RetinaNet \cite{lin2017focal} introduced focal loss for foreground--background imbalance, and FCOS \cite{tian2019fcos} removed anchor boxes. Recent real-time families include YOLOv10 \cite{wang2024yolov10}, YOLOv11 \cite{khanam2024yolov11}, YOLOv12 \cite{tian2025yolov12}, and YOLO26 \cite{jocher2026yolo26}. Transformer detectors began with DETR \cite{carion2020end}; Deformable DETR \cite{zhu2020deformable} improved convergence, and RT-DETR \cite{zhao2024detrs} brought end-to-end transformer detection into the real-time regime.

\textbf{Lightweight detection.} MobileNetV2 \cite{sandler2018mobilenetv2} and MobileNetV3 \cite{howard2019searching} use inverted residual blocks and depthwise separable convolutions to reduce computation, while EfficientDet \cite{tan2020efficientdet} jointly scales backbone, neck, and head. Slicing-based inference such as SAHI \cite{akyon2022slicing} can recover small-object accuracy on high-resolution inputs at the cost of multiple forward passes. SkyDet examines a different endpoint: a 1.22 M parameter, sub-5 MB baseline for mixed land and maritime detection.

\section{The SkySeaLand Dataset}
\label{sec:dataset}

Fig.~\ref{fig1} summarizes the full pipeline from image collection through annotation, verification, format export, and benchmark evaluation.

\subsection{Collection and Annotation}

Images were collected from Google Earth Pro for academic research use. Candidate regions were explored manually over airports, highways, harbors, marinas, and coastal zones in Asia, Europe, Russia, and the United States, then exported at the highest available resolution. Files with heavy noise, strong cloud cover, or duplicate viewpoints were removed during screening, and basic cropping was applied where needed to keep target objects in frame.

Each object instance was annotated with an axis-aligned bounding box and one of four class labels (airplane, boat, car, ship) using CVAT and Roboflow. A second verification pass corrected box placement and class assignment, and invalid or anomalous annotations were removed before the splits were finalized. Final annotations are released in COCO JSON format and in per-image YOLO text format with class indices 0 to 3, so the dataset can be consumed directly by both COCO-style evaluation toolkits and Ultralytics training pipelines without conversion. The dataset is public on Kaggle and Mendeley Data (DOI: 10.17632/d42n3cp86p.3) under a CC BY 4.0 license; the release is intended for academic research, and users should also comply with the terms of the source imagery provider.

\begin{figure}[!t]
\centering
\includegraphics[width=\columnwidth]{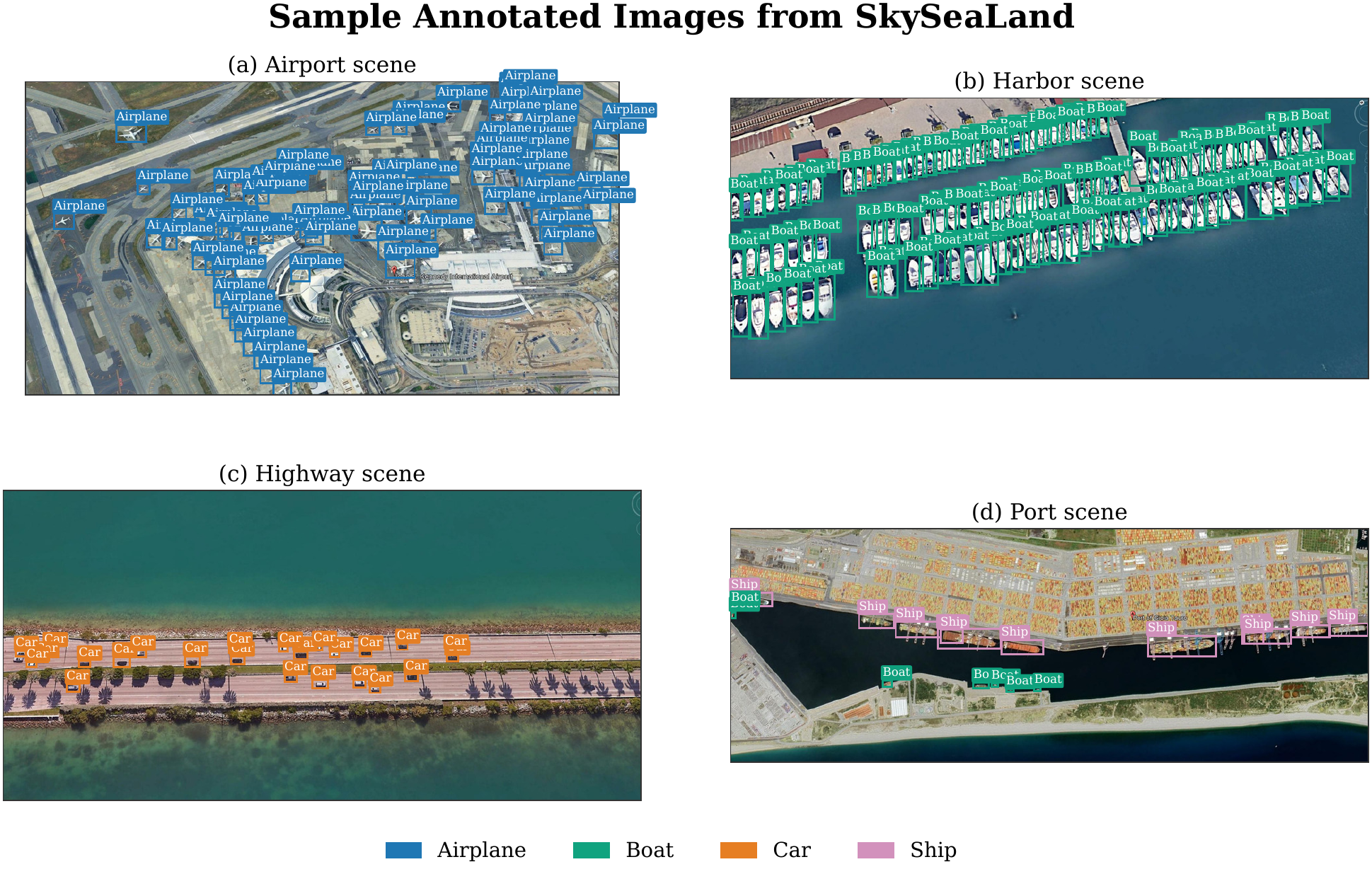}
\caption{Annotated samples from SkySeaLand showing the four object categories: airplane, boat, car, and ship.}
\label{fig2}
\end{figure}

Fig.~\ref{fig2} shows representative airport, harbor, highway, and port scenes after annotation and verification.

\subsection{Splits and Class Distribution}
\label{sec:splits}

The dataset is partitioned into training (1,048 images, 80\%), validation (132 images, 10\%), and test (127 images, 10\%) splits. Table~\ref{tab2} reports image and instance counts per split.

\begin{table}[!t]
\caption{Split Statistics and Per-Class Instance Counts}
\label{tab2}
\centering
\begin{tabular}{lcccc}
\toprule
\textbf{Item}   & \textbf{Train} & \textbf{Val} & \textbf{Test} & \textbf{Total} \\
\midrule
Images          & 1,048 & 132   & 127   & 1,307  \\
Annotations     & 15,034 & 1,992 & 2,075 & 19,101 \\
Airplane        & 3,927 & 367   & 553   & 4,847  \\
Boat            & 2,683 & 657   & 334   & 3,674  \\
Car             & 5,459 & 679   & 798   & 6,936  \\
Ship            & 2,965 & 289   & 390   & 3,644  \\
\bottomrule
\end{tabular}
\end{table}

The car class is the largest, with 6,936 instances (36.3\% of all annotations), followed by airplane (25.4\%), boat (19.2\%), and ship (19.1\%). This distribution arises from the selected scenes and object density: highway and parking images contain more instances than airport or maritime images. Class-weighted focal loss is used during SkyDet training (Section~\ref{sec:skydet}).

Object scale varies widely. The median annotated box covers 2,726 px$^{2}$, the 10th percentile falls below 873 px$^{2}$ (under the COCO small-object threshold of $32^{2}$ px \cite{lin2014microsoft}), and the 90th percentile exceeds 13,000 px$^{2}$. Percentile statistics are consistent across the three splits, which indicates that the splits are scale-balanced.

\subsection{Image Geometry}
\label{sec:geometry}

\begin{figure}[!t]
\centering
\includegraphics[width=\columnwidth]{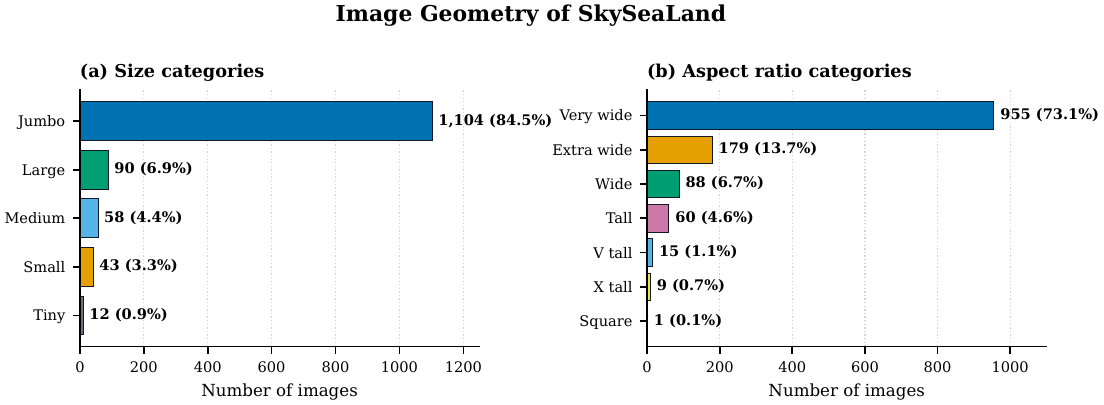}
\caption{Image geometry of SkySeaLand: (a) size categories by longest side; (b) aspect ratio categories. The dataset is dominated by jumbo, very wide scenes.}
\label{fig4}
\end{figure}

SkySeaLand departs from tile-based aerial benchmarks in its raw image geometry. The dataset audit uses five source-size bins (tiny, small, medium, large, and jumbo), with 3,836 px as the boundary of the dominant jumbo group. By longest side, 1,104 images (84.5\%) exceed this boundary, 90 (6.9\%) fall between 1,024 and 3,836 px, 58 (4.4\%) fall between 512 and 1,024 px, and 55 (4.2\%) are below 512 px. Aspect ratios are similarly skewed: 955 images (73.1\%) are near 3:1 width to height, 179 (13.7\%) near 4:1, and 88 (6.7\%) near 2:1, leaving under 7\% in square or tall formats. Cross-tabulation shows the two properties are coupled; 74.6\% of the largest images are near 3:1. Fig.~\ref{fig4} summarizes both distributions.

This profile has direct consequences for detection. Resizing a 3:1 scene to $640 \times 640$ either distorts object shapes or, with letterboxing, maps the image to approximately $640 \times 213$ pixels and leaves about two-thirds of the square tensor as padding. The full horizontal field is retained, but objects are reduced according to the image width and the detector still processes the padded tensor. SkySeaLand therefore provides a practical stress test for wide-scene preprocessing. All trained models use letterbox resizing to preserve aspect ratio.

\section{SkyDet: An Ultra-Lightweight Detection Baseline}
\label{sec:skydet}

\subsection{Design Goals and Architecture}

SkyDet is designed to document the low end of the benchmark's footprint range. Accuracy is traded deliberately for size; the question is how much detection capability remains near one million parameters, not whether the model can match detectors tens of times larger.

The architecture has three components. A MobileNetV3-Small backbone \cite{howard2019searching}, pretrained on ImageNet, uses inverted residual blocks and depthwise separable convolutions \cite{sandler2018mobilenetv2}. A 64-channel feature pyramid \cite{lin2017feature} fuses P3--P5 features. An FCOS-style anchor-free head \cite{tian2019fcos}, also built from depthwise separable convolutions, predicts class scores and boxes without dataset-specific anchor selection. These established components are combined as a compact reference configuration; their individual contributions are not isolated in this study.

\subsection{Training}

Training uses class-weighted focal loss \cite{lin2017focal}, with weights inversely proportional to class frequency, and a cosine learning-rate schedule \cite{loshchilov2016sgdr} initialized at $3 \times 10^{-4}$. Inputs use letterbox resizing. For validation diagnostics, the score threshold increases from 0.001 to 0.01 at epoch 10 and to 0.05 at epoch 20; this schedule affects evaluation and checkpoint monitoring, not gradient updates. Final test inference uses 0.05. SkyDet was trained for 150 epochs on a Tesla T4, and the best validation checkpoint occurred at epoch 124. Table~\ref{tab3} summarizes the configuration.

\begin{table}[!t]
\caption{SkyDet Configuration}
\label{tab3}
\centering
\footnotesize
\begin{tabular}{@{}p{0.28\columnwidth}p{0.64\columnwidth}@{}}
\toprule
\textbf{Component}              & \textbf{Configuration} \\
\midrule
Backbone                        & MobileNetV3-Small, ImageNet pretrained \\
Feature fusion                  & FPN, 64-channel width \\
Detection head                  & Anchor-free, depthwise separable \\
Loss                            & Class-weighted focal loss \\
LR schedule                     & Cosine decay, initial $3 \times 10^{-4}$ \\
Validation threshold            & 0.001 to 0.05 by epoch 20 \\
Epochs / best checkpoint        & 150 / 124 \\
Parameters / checkpoint size    & 1.22 M / 4.90 MB \\
\bottomrule
\end{tabular}
\end{table}

\begin{table*}[!t]
\caption{Detector Performance on the SkySeaLand Test Split, Sorted by mAP50-95. $^\dagger$Constrained 1,000-iteration run.}
\label{tab4}
\centering
\begin{tabular}{lcccccc}
\toprule
\textbf{Model} & \textbf{mAP50 (\%)} & \textbf{mAP50-95 (\%)} & \textbf{Params (M)} & \textbf{Size (MB)} & \textbf{Latency (ms)} & \textbf{GPU} \\
\midrule
RT-DETR-x \cite{zhao2024detrs} & 87.32 & 60.36 & 65.48 & 131.2 & 65.70 & Tesla T4 \\
YOLO26x \cite{jocher2026yolo26} & 88.20 & 60.00 & 55.64 & 118.3 & 19.20 & Tesla T4 \\
RT-DETR-l \cite{zhao2024detrs} & 87.60 & 59.90 & 31.99 & 66.2 & 8.90 & NVIDIA L4 \\
YOLOv11m \cite{khanam2024yolov11} & 87.20 & 59.00 & 20.03 & 40.5 & 7.30 & Tesla T4 \\
YOLO26l \cite{jocher2026yolo26} & 87.20 & 58.60 & 24.75 & 53.0 & 8.60 & Tesla T4 \\
YOLOv11x \cite{khanam2024yolov11} & 87.00 & 58.60 & 56.83 & 114.4 & 2.70 & NVIDIA A100 \\
YOLOv12x \cite{tian2025yolov12} & 87.00 & 58.50 & 59.05 & 119.1 & 4.10 & NVIDIA A100 \\
YOLOv12m \cite{tian2025yolov12} & 87.10 & 57.80 & 20.11 & 40.8 & 13.30 & Tesla T4 \\
YOLOv10m \cite{wang2024yolov10} & 84.40 & 56.40 & 15.32 & 33.5 & 8.30 & Tesla T4 \\
Faster R-CNN$^\dagger$ \cite{ren2015faster} & 46.15 & 28.40 & 41.75 & 283.9 & 95.20 & Tesla T4 \\
DETR ResNet-50 \cite{carion2020end} & 60.80 & 26.70 & 41.50 & 166.0 & 47.55 & Tesla T4 \\
SkyDet (ours) & 60.50 & 24.32 & 1.22 & 4.90 & 13.74 & Tesla T4 \\
\bottomrule
\end{tabular}
\end{table*}

\section{Experiments on SkySeaLand}
\label{sec:experiments}

\subsection{Setup}

We benchmark twelve detectors on SkySeaLand: SkyDet; RT-DETR-l and RT-DETR-x \cite{zhao2024detrs}; YOLOv10m \cite{wang2024yolov10}; YOLOv11m and YOLOv11x \cite{khanam2024yolov11}; YOLOv12m and YOLOv12x \cite{tian2025yolov12}; YOLO26l and YOLO26x \cite{jocher2026yolo26}; DETR \cite{carion2020end}; and Faster R-CNN with a ResNet-50 FPN backbone. All models use the split in Section~\ref{sec:splits} and $640 \times 640$ letterboxed inputs. Evaluation follows COCO metrics \cite{lin2014microsoft}, reporting mAP over IoU thresholds 0.50--0.95 (mAP50-95) and at IoU 0.50 (mAP50) on the test split. SkyDet uses a final score threshold of 0.05, non-maximum suppression at 0.45 IoU, and at most 300 detections per image.

Models use fixed, model-specific recipes: 100 epochs for YOLO and RT-DETR, 150 for SkyDet, 50 for DETR, and 1,000 iterations for Faster R-CNN. These unequal budgets make the accuracy table a practical descriptive benchmark rather than a controlled architecture study. Hardware also varies across runs (Tesla T4, NVIDIA L4, and NVIDIA A100). Parameter counts and measured checkpoint sizes provide footprint context, while latency comparisons are restricted to rows measured on the same GPU.

SkyDet checkpoint selection used only the validation split. The selected checkpoint was then evaluated on the held-out test split, producing 60.50 mAP50 and 24.32 mAP50-95. Its mean Tesla T4 inference time was recorded as 0.013741 s per image and converted to 13.74 ms (72.8 FPS). The reported 4.90 MB is the serialized checkpoint size measured on disk, rather than a value inferred from parameter count.

\subsection{Results}

Three observations follow from Table~\ref{tab4}. First, the tested YOLO and RT-DETR variants occupy a narrow descriptive band of 84.4--88.2 mAP50 and 56.4--60.4 mAP50-95 despite ranging from 15 M to 65 M parameters. Larger parameter counts do not yield consistent gains among these runs; for example, YOLOv11m exceeds YOLOv11x by 0.4 mAP50-95 points with 35\% of the parameters. The single-run, model-specific recipes preclude statistical equivalence claims. Second, the 50-epoch DETR result is a constrained-budget reference because DETR variants commonly require longer schedules \cite{zhu2020deformable}. The daggered Faster R-CNN row is retained for completeness but excluded from family-level interpretation because its 1,000-iteration budget is substantially shorter than the other training schedules.

\begin{figure}[!t]
\centering
\includegraphics[width=\columnwidth]{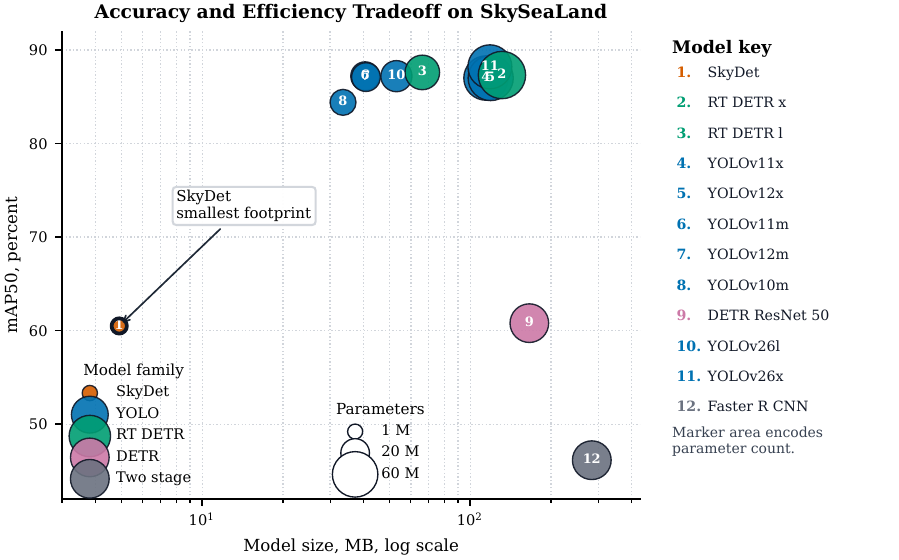}
\caption{Accuracy versus model size on SkySeaLand (log scale). The modern detectors cluster in a narrow accuracy band across a 4x size range, while SkyDet defines a separate low-footprint operating point.}
\label{fig5}
\end{figure}

Third, SkyDet occupies the table's low-footprint endpoint. At 1.22 M parameters and 4.90 MB, it has about 13$\times$ fewer parameters than YOLOv10m and a 27$\times$ smaller checkpoint than RT-DETR-x. It retains 60.5 mAP50, or 69\% of YOLOv11m's mAP50, with 6\% of its parameters. Compared with the constrained DETR run, SkyDet is within 2.4 mAP50-95 points and 0.3 mAP50 points at about 1/34 of the checkpoint size. On the same Tesla T4, its 13.74 ms latency (72.8 FPS) is 21\% of RT-DETR-x's 65.70 ms. These comparisons establish a footprint--accuracy reference point, not an accuracy advantage. Fig.~\ref{fig5} visualizes the separation in accuracy--checkpoint-size space.

Parameter count, storage, and runtime should not be conflated. SkyDet is the smallest model in Table~\ref{tab4}, but its T4 latency is higher than the reported YOLOv11m, YOLOv10m, YOLO26l, and YOLOv12m runs. Operator choice, memory access, framework kernels, and execution overhead can outweigh parameter count at this scale. The defensible deployment result is therefore a 4.90 MB storage footprint with measured real-time throughput; the paper claims a latency advantage only over RT-DETR-x on the same hardware.

\section{Cross-Dataset Evaluation}
\label{sec:cross}

Table~\ref{tab5} reports secondary diagnostics on DOTA v2.0 and RSOD. Models are trained independently on each dataset without weight transfer. The available SkyDet runs use dataset-specific configurations that are not parameter-identical, comparator coverage is asymmetric, and comparator training budgets are not established as matched in the available record. Consequently, this experiment supports neither a common-footprint claim nor cross-domain generalization of one checkpoint. It asks only whether the training and evaluation pipeline can be applied beyond SkySeaLand. The DOTA diagnostic uses a 1,822-image Roboflow-exported subset of DOTA v2.0 \cite{xia2018dota} retaining all 18 categories in COCO horizontal-box format; it is not the full official release. RSOD \cite{long2017accurate} uses all four categories. SkyDet latency was measured with two Tesla T4 GPUs, whereas comparator latency used one T4; the timings are reported only for provenance and are not compared across rows.

\begin{table}[!t]
\caption{Selected Cross-Dataset Diagnostics with Hardware Provenance}
\label{tab5}
\centering
\scriptsize
\setlength{\tabcolsep}{2pt}
\begin{tabular}{@{}llrrrc@{}}
\toprule
\textbf{Dataset} &
\textbf{Model} &
\textbf{mAP50 (\%)} &
\textbf{mAP50-95 (\%)} &
\textbf{Latency (ms)} &
\textbf{GPU(s)} \\
\midrule
DOTA subset & SkyDet & 14.70 & 6.70 & 11.59 & $2\times$T4 \\
DOTA subset & Faster R-CNN & 15.16 & 7.35 & 87.97 & $1\times$T4 \\
RSOD & SkyDet & 88.90 & 54.30 & 11.18 & $2\times$T4 \\
RSOD & YOLOv12x & 95.60 & 72.00 & 46.10 & $1\times$T4 \\
\bottomrule
\end{tabular}
\end{table}

On the DOTA v2.0 subset, SkyDet is within 0.46 mAP50 points and 0.65 mAP50-95 points of Faster R-CNN, while both absolute scores remain low on the horizontal-box export. On RSOD, SkyDet trails YOLOv12x by 6.7 mAP50 points and 17.7 mAP50-95 points. These accuracy differences are descriptive because the training budgets are not established as matched. The selected comparisons show that the pipeline produces measurable results on two additional aerial datasets, but the asymmetric baselines and dataset-specific SkyDet configurations do not establish general superiority, a transferable efficiency ratio, or a latency advantage.

\section{Discussion and Limitations}
\label{sec:limitations}

SkyDet's 36.18-point gap between mAP50 and mAP50-95 indicates that performance declines substantially as localization criteria become stricter. This pattern is consistent with the sensitivity of small bounding boxes to pixel-level displacement, but the present evidence cannot assign the gap to object scale: no current-checkpoint per-scale or per-class AP is available. The result should therefore motivate, rather than validate, future tests of tiling, higher input resolution, and scale-aware training.

The conclusions are bounded in five ways. First, SkySeaLand is smaller than DOTA \cite{xia2018dota} and DIOR \cite{li2020object}; it is a compact complementary benchmark. Duplicate views were screened before splitting, but the split is image-level and does not enforce geographic separation, so regional similarity may remain across subsets. The aggregate evaluation also does not report land-versus-maritime performance separately. Second, the detector table contains single runs under unequal model-specific recipes, and no uncertainty estimates support claims about small differences. Third, mixed hardware and device counts prevent most latency comparisons; the sole stated ratio uses the single-T4 SkySeaLand runs for SkyDet and RT-DETR-x, while external dual- and single-T4 timings are not compared. Fourth, no component ablation or current-checkpoint per-scale AP is reported, so the study makes no causal claim about SkyDet's errors. Finally, the external-dataset comparison uses asymmetric baselines and non-identical SkyDet configurations. SkySeaLand annotations are axis-aligned, so oriented detection is outside the present scope.

\section{Conclusion}

SkySeaLand contributes 1,307 public satellite images and 19,101 annotations across four land and maritime transportation classes, with native COCO and YOLO labels and documented wide-scene geometry. Its twelve-detector benchmark places the tested YOLO and RT-DETR runs in a narrow descriptive accuracy range. SkyDet defines the low-footprint endpoint, attaining 60.5 mAP50 and 24.32 mAP50-95 with 1.22 M parameters, a 4.90 MB checkpoint, and 13.74 ms latency (72.8 FPS) on a Tesla T4. It quantifies accuracy retained under a 5 MB budget rather than challenging high-accuracy detectors. Future work should evaluate geographically separated splits, repeated runs, and small-object interventions.

\section*{Data Availability}

SkySeaLand is publicly available through Mendeley Data at \href{https://doi.org/10.17632/d42n3cp86p.3}{doi:10.17632/d42n3cp86p.3} and through \href{https://www.kaggle.com/datasets/mdzahidhasanriad/skysealand}{Kaggle}. The release contains all 1,307 images, the fixed train/validation/test subsets, and four-class annotations in COCO and YOLO formats prepared with CVAT and Roboflow.

\balance
\bibliography{references}

\end{document}